\documentclass{article}
\usepackage{spconf,amsmath,graphicx}

\usepackage[normalem]{ulem}
\usepackage{bm}
\usepackage{dsfont}
\usepackage{diagbox}
\usepackage{array}
\usepackage{multirow}
\usepackage{booktabs}
\usepackage{nccmath}
\usepackage{caption}
\usepackage{makecell}
\usepackage{subcaption}
\usepackage{float} 
\usepackage[dvipsnames]{xcolor}
\usepackage{ulem}
\usepackage{cite}
\usepackage{url}

\usepackage{amsmath}
\usepackage{algpseudocode}
\usepackage{amsthm}
\usepackage{dsfont}
\usepackage{algorithm}
\usepackage{thmtools}
\usepackage{amssymb}
\usepackage[dvipsnames]{xcolor}

\def\bn{\boldsymbol{n}}

\def\bx{\boldsymbol{x}}
\def\by{\boldsymbol{y}}

\def\bH{\boldsymbol{H}}

\def\bW{\boldsymbol{W}}

\def\bUpsilon{\boldsymbol{\Upsilon}}

\def\calL{\mathcal{L}}

\def\calU{\mathcal{U}}

\newcommand{\probP}[1]{\mathds{P}\mathrm{r}\left\{{#1}\right\}}

\def\eg{\emph{e.g.}}
\def\ie{\emph{i.e.}}
\def\wrt{\emph{w.r.t.~}}
\def\etal{\emph{et al.}}

\def\conv{\circledast}

\theoremstyle{definition}

\algnewcommand\algorithmicinput{\textbf{Input:}}
\algnewcommand\Input{\item[\algorithmicinput]}
\algnewcommand\algorithmicoutput{\textbf{Output:}}
\algnewcommand\Output{\item[\algorithmicoutput]}
\algnewcommand\algorithmicinit{\textbf{Initialize:}}
\algnewcommand\Init{\item[\algorithmicinit]}

\def\To{\textbf{to}~}

\newcommand{\subsec}[1]{\noindent{\textbf{#1~~}}}
\title{Deep-URL: A MODEL-AWARE APPROACH TO BLIND DECONVOLUTION BASED ON DEEP UNFOLDED RICHARDSON-LUCY NETWORK}
\name{Chirag~Agarwal, Shahin~Khobahi, Arindam~Bose, Mojtaba~Soltanalian, Dan~Schonfeld}
\address{Department of Electrical \& Computer Engineering, University of Illinois at Chicago, Chicago, IL 60607.}
\begin{document}
%
\maketitle
\begin{abstract}
    The lack of interpretability in current deep learning models causes serious concerns as they are extensively used for various life-critical applications. 
    Hence, it is of paramount importance to develop interpretable deep learning models.
    In this paper, we consider the problem of blind deconvolution and propose a novel model-aware deep architecture that allows for the recovery of both the blur kernel and the sharp image from the blurred image.
    In particular, we propose the Deep Unfolded Richardson-Lucy (Deep-URL) framework --- an interpretable deep-learning architecture that can be seen as an amalgamation of classical estimation technique and deep neural network, and consequently leads to improved performance. 
    Our numerical investigations demonstrate significant improvement compared to state-of-the-art algorithms.
\end{abstract}
\begin{keywords}
Blind deconvolution, model-aware deep learning, machine learning, deep unfolding, non-convex optimization
\end{keywords}
\section{Introduction}
\label{sec:intro}
In digital photography, motion blur is a common and longstanding problem where the blurring is induced by the relative motion of the camera or the subject with respect to the other \cite{lai2016comparative}.
In classical image processing, such a motion blur is generally regarded as a motion kernel being applied on the original sharp image through a linear operation, \eg, convolution.
Often in practice, however, neither the blur kernel nor the original image is known \textit{a priori}, and thus the task becomes to estimate both from the blurry input image.
In image processing, the term \textit{blind deconvolution} is often used to represent the task of image restoration without any explicit knowledge of the impulse response function, also known as the point-spread function (PSF) and the original sharp image \cite{lai2016comparative, levin2009understanding}.
The blurred image $\by$ is typically formulated as:
{
    \begin{align}
        \by = \bH \conv \bx + \bn,
    \end{align}
}
where $\bx$ and $\bH$ are the unknown original clean image and the blur kernel, respectively, $\bn$ is the additive measurement noise generally modeled as white Gaussian noise (AWGN) with variance $\sigma^2$, and $\conv$ represents the 2D convolution operator.
Hence, the task of blind deconvolution is to estimate a sharp $\bx$ and the corresponding $\bH$ from an infinite set of pairs $(\bx, \bH)$ using the blurry image $\by$, making it an ill-posed and very challenging problem.

A judicious approach to such problems is to utilize some prior knowledge about the statistics of the natural image and/or motion kernels. There exists a multitude of algorithms to efficiently estimate the image $\bx$ and kernel $\bH$ using prior knowledge of the model \cite{fergus2006removing, levin2006blind, xu2010two}.
A majority of them are based on maximum-a-posterior (MAP) framework,
{
    \begin{align}
        (\hat{\bx}, \hat{\bH}) &= \arg \max_{\bx, \bH} {\probP {\bx, \bH~|~\by}}, \nonumber\\
        &= \arg \max_{\bx, \bH} {\probP {\by~|~\bx, \bH} \probP{\bx} \probP{\bH}},
        \label{eq:map}
    \end{align}
}
where $\probP{\by~|~\bx, \bH}$ is the likelihood of the noisy output $\by$ given a certain $(\bx, \bH)$, that corresponds to the data fidelity term, and $\probP{\bx}$ and $\probP{\bH}$ are the priors of the original image and blur kernel, respectively.
Note that, Eq.~(\ref{eq:map}) is correct under the assumption that the sharp original image $\bx$ and the blur kernel $\bH$ are independent.  
These MAP-based algorithms are often \textit{iterative} in nature and usually rely on the sparsity-inducing regularizers, either in gradient domain \cite{xu2010two, krishnan2011blind, xu2013unnatural} 
or more generally in sparsifying transformation domain
\cite{pan2018deblurring}.
However, the knowledge of the prior is not usually enough, for instance, Levin \etal \cite{levin2009understanding} shows that MAP-based methods may lead to a trivial solution of an impulse kernel resulting in the same noisy image as output.
By carefully designing the appropriate regularizer and selecting the proper step size and learning rate, one may find a sharper image.
These parameters are, however, difficult to determine analytically as they heavily depend on the noisy input image itself, and thus do not admit any generalization.

Data-driven methods, on the other hand, make an attempt to determine a non-linear mapping that deblurs the noisy image by learning the appropriate parameter choices particular to an underlying image dataset using deep neural networks (DNN) \cite{xu2018motion, chakrabarti2016neural}.
Given the training dataset, one can use a DNN  either to extract features from the noisy image to estimate the blur kernel \cite{chakrabarti2016neural} 
or directly learn the mapping to the sharp image \cite{nah2017deep}.
Although these methods achieve substantial performances in certain practical scenarios, they often do not succeed in handling various complex and large-sized blur kernels in blind deconvolution.
The structure of the neural networks is usually empirically determined and thus they often lack inherent interpretability.
Recent works generate attribution based maps to explain the networks decision, however, they disregard the untapped potential of the model knowledge \cite{agarwal2019removing}.

\begin{figure}[t]
    \centering
    \includegraphics[scale=0.45]{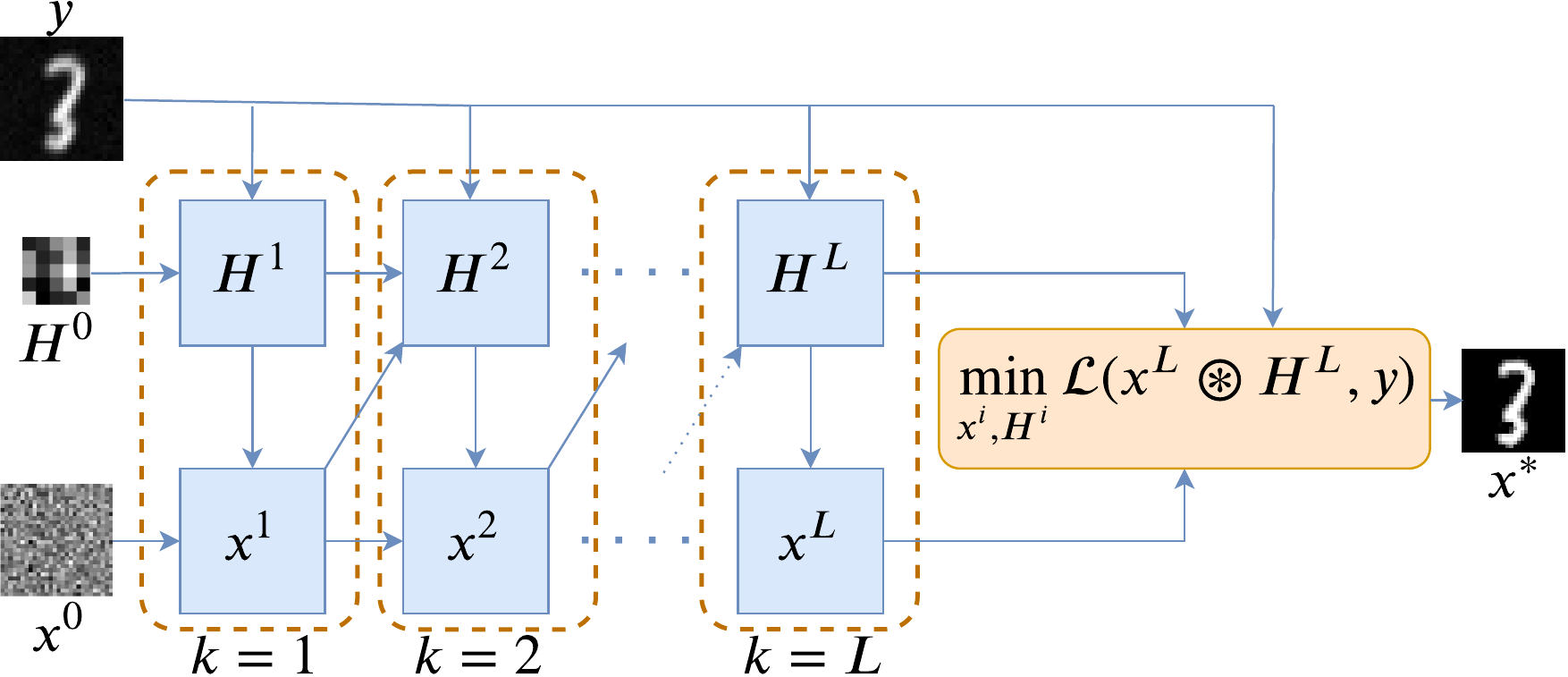}
    \caption{\small
        Proposed Deep-URL architecture for model-aware blind deconvolution. 
        Given a blurred image $\by$ and initial estimates of the clean image $\bx^{0}$ and blurring kernel $\bH^{0}$, the model updates $\bx^{k}$ and $\bH^{k}$ 
        following Algorithm~\ref{alg:DURL}.
        }
    \label{fig:model_arch}
    \vspace{-0.6cm}
\end{figure}
\noindent In order to enjoy the advantages of both model-based iterative algorithms and data-driven learning strategies, one may exploit the idea of \textit{deep unfolding} \cite{hershey2014deep, gregor2010learning}.
Especially, \cite{hershey2014deep} shows that propagating through a neural network is, in fact, \emph{equivalent} to implementing the iterative algorithm for a finite number of times, and thus the trained network can be naturally interpreted as a parameter optimized algorithm.
In recent years, deep unfolding networks have gained a significant amount of attention in various branches of signal processing \cite{hershey2014deep, khobahi2019deepsignal, khobahi2019deepradar, bertocchi2019deep,khobahi2019model}.
However, in the context of blind image deconvolution, the extent of deep unfolding capabilities remains largely unexplored.
Recently Li \etal \cite{li2019algorithm}, performed motion deblurring by means of unfolding an iterative algorithm that relies on total-variation (TV) regularization prior in the image gradient domain \cite{perrone2016clearer}.
Although this approach performs better than the state-of-the-art model-based and data-driven blind deconvolution counterparts, the strict requirement of training a network for a certain dataset makes the algorithm impractical for real-time usage: the algorithm requires a ground truth dataset, to begin with.
Additionally, in practice, motion kernels are not pre-deterministic (\eg, in drone image processing), and hence acquiring a labeled dataset is not possible for a supervised learning scenario.

In this paper, we propose a novel technique to unfold an iterative algorithm that estimates the latent clean image and corresponding blur kernel \textit{on the fly} --- a zero-shot self-supervised algorithm.
In particular, we use the classical Richardson-Lucy blind deconvolution algorithm \cite{fish1995blind} to construct the network structure and iteratively estimate the clean image and the kernel.
We experimentally verify the performance of our algorithm and compare it with \cite{li2019algorithm} and other iterative algorithms and recent neural network approaches.

\section{Problem Formulation}\label{sec:richardson}
In this section, we lay the groundwork for our proposed model-aware deep architecture for the problem of blind deconvolution. To this end, we consider an extension of the Richardson-Lucy (RL) algorithm as a baseline to design a deep neural network such that each layer imitates the behavior of one iteration of the RL algorithm.

Generally, the problem of blind deconvolution can be cast as the following optimization problem:
{
    \begin{align}\label{eq:form}
        \min_{\bx,\bH}\; \|\by - \bH\conv \bx\|_2^2 + \lambda\mathrm{TV}(\bx),
    \end{align}
}
where the first term represents the data fidelity term and $\lambda$ is the regularization coefficient for the total variation (TV) regularization operated on the image $\bx$.
The RL algorithm seeks to recover the sharp image $\bx$ and the blur kernel $\bH$ in an iterative manner as described in \cite{fish1995blind}.
Starting from an initial guess for the sharp image and the kernel $(\bx^0, \bH^0)$, the update steps for the image and the kernel at the $k$-th iteration is given by,
{
\begin{subequations}
	\begin{align}
	    \bH^{k+1} &= \left(\left[ \frac{\by}{\bx^k \conv \bH^k}\right] \conv {\bx^k}^{\dagger} \right) \odot \bH^k, \label{eq:rl_h_update}\\
		\bx^{k+1} &= \left(\left[ \frac{\by}{\bx^k \conv \bH^{k+1}}\right] \conv {\bH^{k+1}}^{\dagger} \right) \odot \bx^k, \label{eq:rl_x_update} 
	\end{align}
\end{subequations}
}
where $\odot$ represents the Hadamard product and $(\cdot)^{\dagger}$ denotes the flipped version of the vector/matrix argument.
\section{Blind Deconvolution via Deep-URL}
\label{sec:method}
In order to obtain a model-aware deep architecture we slightly over parameterize the iterations of RL algorithm (See Eq.~(\ref{eq:rl_h_update})-(\ref{eq:rl_x_update})) and unfold them onto the layers of a deep neural network.
In particular, each layer corresponds to one iteration of the baseline iterative algorithm. Namely, we fix the total computational complexity of the RL algorithm by fixing the total number of iterations as a DNN with $L$ layers.
Thus, by substituting the $\bx^{k}$ and $\bH^{k}$ in Eq.~(\ref{eq:rl_h_update})-(\ref{eq:rl_x_update}) with trainable parameters, we reformulate each subsequent iterative operation as:

{
    \footnotesize
    \begin{subequations}
        	\begin{align}
        		\bH^{k+1} &= \sigma\left(\mathrm{ReLU}\left(\left[ \frac{\by}{\mathrm{ReLU}(\bx^k \conv \bW_{\bH}^{k})}\right] \conv {\bx^k}^{\dagger} \right) \odot \bW_{\bH}^{k}\right), \label{eq:unfold_h_update}\\
        		\bx^{k+1} &= \sigma\left(\mathrm{ReLU}\left(\left[ \frac{\by}{\mathrm{ReLU}(\bW_{\bx}^{k} \conv \bH^{k+1})}\right] \conv {\bH^{k+1}}^{\dagger} \right) \odot \bW_{\bx}^{k}\right),
        		\label{eq:unfold_x_update}
        	\end{align}
    \end{subequations}
}%
where $\bW^{k}_{\bx}$ and $\bW^{k}_{\bH}$ are the weights for $k$-th layer. Furthermore, $\sigma(\cdot)$ represents the \emph{Sigmoid} activation function and $\mathrm{ReLU}$ denotes the Rectifier Linear Unit.
Note that there exist two implicit constraints on the recovered sharp image and the kernel: (a) both $\bx$ and $\bH$ are non-negative and (b) each element of $\bx$ and $\bH$ must meet a range constraint.
Hence, in order to ensure constraint (a), each convolution operation is activated by a $\mathrm{ReLU}$ function, and in addition we use the \emph{Sigmoid} activation after each update step to satisfy constraint (b).

Let $\bUpsilon^{k} = \{\bW^{k}_{\bx}, \bW^{k}_{\bH}\}$ denote the set of trainable parameters of layer $k$, and $\bUpsilon = \bUpsilon^1 \cup \bUpsilon^2 \cup \cdots \cup \bUpsilon^L$.
Using the iterative updates from Eq.~(\ref{eq:unfold_h_update})-(\ref{eq:unfold_x_update}), we formulate the training of our proposed model-aware deep network: \textit{Deep Unfolded Richardson Lucy} (Deep-URL) architecture as follows,
{
\begin{equation}
    \min_{\bUpsilon} \calL( \bx^{L}\conv \bH^{L}, \by) + \lambda \mathrm{TV}(\bx^{L})
    \label{eq:optimization}
\end{equation}
}
where the loss function $\calL(\cdot)$ is the negative of the structural similarity index (SSIM) \cite{wang2004image} between the true blurred image $\by$ and the reconstructed blurred image $\hat{\by} = \bx^{L}\conv \bH^{L}$.

It is worth mentioning that the proposed deep architecture in conjunction with the proposed learning method manifests itself as a \textit{self-supervised} learning process where the degraded image $\by$ is the only information used for estimation of the sharp image $\bx^{*}$ and the blurred kernel $\bH^{*}$.
Fig.~\ref{fig:model_arch} illustrates the proposed Deep-URL architecture and the training process. 
Finally, Algorithm~\ref{alg:DURL} summarizes the joint optimization process for updating $\bUpsilon$.
Note that, \emph{once the self-supervised model is optimized for a given blur kernel, the learned weights can be directly used for deblurring any image blurred with the same kernel.}
\begin{algorithm}[ht]
    \small
	\caption{\textsc{Deep-URL}}
	\begin{algorithmic}[1]
		\Input{$\by$: blurred image, $L$: number of layers, $N$: number of epochs}
		\Output{$\bH^{*}$: estimated kernel, $\bx^{*}$: sharp image}
		\Init{$\bH^{0}\gets~\calU(0,1)$;
		$\bx^{0}\gets~\calU(0,1)$}
		\For{$i=1$ \To $N$}
    		\For{$k=0$ \To $L-1$}
                \State{Compute $\bH^{k+1}$ using Eq.~(\ref{eq:unfold_h_update}).}
                \State{Compute $\bx^{k+1}$ using Eq.~(\ref{eq:unfold_x_update}).}
    		\EndFor
    		\State{Compute the gradients of Eq.~(\ref{eq:optimization}) \wrt $\bUpsilon$.}
    		\State{Update $\bUpsilon$.}
    		\State{$\bH^{0} \gets \bH^{L}; \bx^{0} \gets \bx^{L}$}
		\EndFor
		\State{$\bH^{*} = \bH^{L}; \bx^{*} = \bx^{L}$}
	\end{algorithmic}
	\label{alg:DURL}
\end{algorithm}
\vspace{-0.2cm}
\vspace{-0.4cm}
\section{Experiments}
\label{sec:exp}
In this section, we investigate the performance of the proposed Deep-URL framework and compare it with several other state-of-the-art methods in the context of blind deconvolution.
First, we compare the performance of Deep-URL with the baseline RL algorithm using the standard MNIST handwritten digit dataset \cite{lecun1998gradient}.
Second, we use Levin dataset \cite{levin2009understanding} to compare Deep-URL with existing iterative and deep learning-based blind deconvolution methods proposed in \cite{li2019algorithm, chakrabarti2016neural, nah2017deep}.

\subsec{Optimization setup.}
The training of Deep-URL (Eq.~(\ref{eq:optimization})) is carried out using the RMSprop optimizer for $5000$ epochs by employing an adaptive learning rate scheme with an initial learning rate of $0.1$ and a decaying factor of $0.1$ when reaching $40$\% and $60$\% of the total number of epochs. In addition, the TV regularization coefficient $\lambda$ was set to $0.1$ for all experiments.
All trainable parameters were initialized using a uniform distribution.
We performed a batch-wise optimization, with a batch size of $4$, on images blurred using the same kernel for enhancing the performance of Deep-URL. 

\subsec{Evaluation metrics.}
Inspired by \cite{li2019algorithm}, we use the following metrics to evaluate the performance of our proposed method: (1) Structural Similarity Index (SSIM), (2) Peak Signal-to-Noise-Ratio (PSNR), (3) Improvement in Signal-to-Noise-Ratio (ISNR) for the quality of the reconstructed image $\bx^*$, and (4) Root-Mean-Square Error (RMSE) for comparing the recovered blur kernel $\bH^*$ with the original $\bH$. In the sequel, we use the term PSF and blur kernel interchangeably.
\begin{table}[]
\small
\centering
\caption{
    \small
    Evaluation metric scores averaged over 1000 MNIST images.
    Across all image quality metrics, Deep-URL (D-URL) outperforms the RL algorithm.
}
\label{tab:MNIST}
\vspace{-0.2cm}
\begin{tabular}{@{}l|rr|rr@{}}
\toprule
&\multicolumn{2}{c|}{$L=2$} & \multicolumn{2}{c}{$L=5$}\\
\cline{2-5}
Metrics & \multicolumn{1}{c}{RL} & \multicolumn{1}{c}{D-URL} & \multicolumn{1}{|c}{RL} & \multicolumn{1}{c}{D-URL} \\ \midrule
PSNR(dB) & 10.3919 & \textbf{18.2821} & 10.4742 & \textbf{19.7075} \\
ISNR (dB) & 0.0651 & \textbf{7.9554} & 0.0764 & \textbf{9.3096} \\
SSIM & 0.4453 & \textbf{0.7669} & 0.4484 & \textbf{0.8206} \\
RMSE($\times$1e-3) & 38.54 & \textbf{4.396} & 38.07 & \textbf{4.399} \\
\bottomrule
\end{tabular}
\vspace{-0.3cm}
\end{table}
\begin{figure}[h]
    \centering
    \vspace{-0.2cm}
    \includegraphics[width=0.45\textwidth]{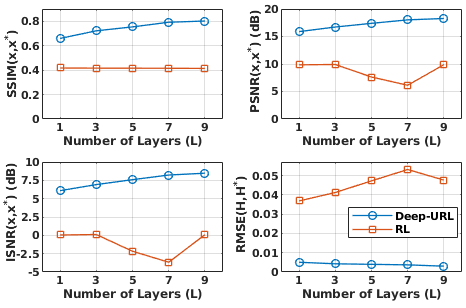}
    \vspace{-0.3cm}
    \caption{
        \small
        The evaluation metric scores across all image and kernel for different number of layers ($L$) show that the performance of Deep-URL increases on increasing $L$ as compared to the baseline RL algorithm.
        }
     \label{fig:layer}
     \vspace{-0.3cm}
\end{figure}
\begin{figure*}[ht]
    \centering
    {
		\small
		\vspace{-0.3cm}
		\begin{flushleft}
		    \hspace{2.6cm} (a)
			\hspace{1.9cm} (b) 
			\hspace{1.9cm} (c)
			\hspace{1.9cm} (d)
			\hspace{1.9cm} (e)
			\hspace{1.9cm} (f)
		\end{flushleft}
		\vspace{-0.3cm}
	}    
    \includegraphics[width=0.8\textwidth]{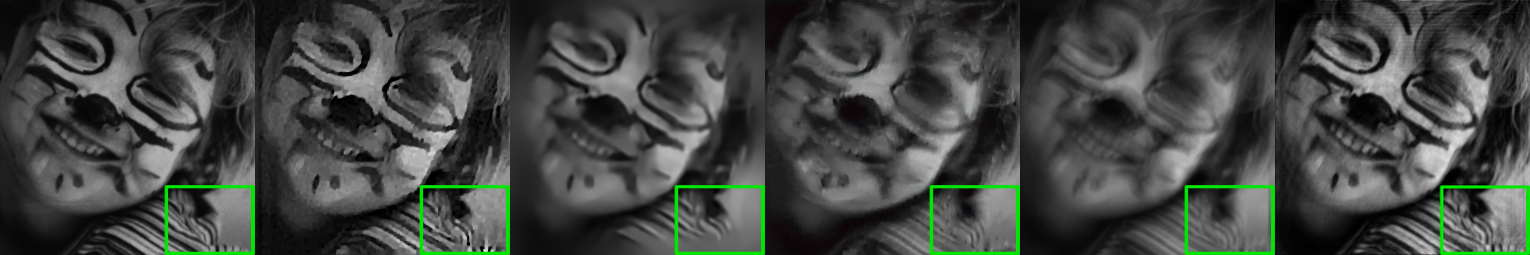}
    {
		\small
		\vspace{-0.3cm}
		\begin{flushleft}
		    \hspace{2.2cm} SSIM: 1.0
			\hspace{0.8cm} 0.6809 \cite{li2019algorithm}
			\hspace{0.8cm} 0.7584 \cite{chakrabarti2016neural}
			\hspace{0.7cm} 0.6082 \cite{nah2017deep}
			\hspace{0.7cm} 0.5315 (RL)
			\hspace{0.2cm} 0.9189 (Deep-URL)
			\vspace{-0.2cm}
		\end{flushleft}
	}
    \includegraphics[width=0.8\textwidth]{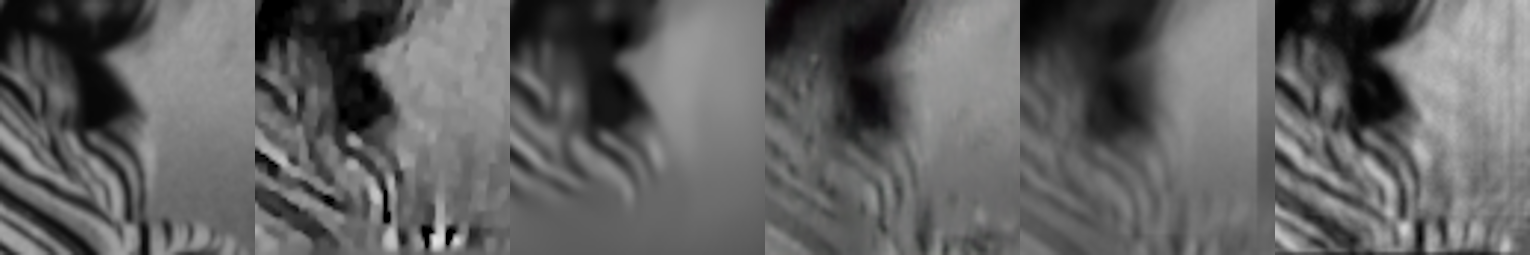}
    \caption{
    \small
    Qualitative results for a sample image from Levin dataset \cite{levin2009understanding} taken from \cite{li2019algorithm}.
    The SSIM score between the ground truth image (row 1, a) and the reconstructed images using different iterative and deep learning based blind deconvolution methods (row 1, b-e) shows the superior performance of Deep-URL (f).
    Comparing the inset images (green boxes in row 1), shows the effectiveness of Deep-URL in retaining fine details of the image.
    Interestingly, the SSIM of \cite{li2019algorithm} (row 1, b) is low since the image is slightly shifted as it fails to reconstruct the blurring kernel correctly (Fig.~\ref{fig:kernel}, b)
    }
    \label{fig:qual}
    \vspace{-0.6cm}
\end{figure*}
\begin{figure}[ht]
    \centering
    {
		\small
		\begin{flushleft}
		    \hspace{1.2cm} (a)
			\hspace{0.9cm} (b) 
			\hspace{0.9cm} (c)
			\hspace{0.9cm} (d)
			\hspace{0.9cm} (e)
		\end{flushleft}
		\vspace{-0.35cm}
	} 
    \includegraphics[width=0.4\textwidth]{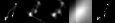}
    {
		\small
		\begin{flushleft}
			\vspace{-0.3cm}
		    \hspace{0.5cm} Ground~Truth
			\hspace{0.15cm} \cite{li2019algorithm} 
			\hspace{0.8cm} \cite{chakrabarti2016neural}
			\hspace{0.8cm} RL
			\hspace{0.4cm} Deep-URL
		\end{flushleft}
		\vspace{-0.3cm}
	}
	\vspace{-0.1cm}
    \caption{
        \small
        Ground truth blur kernel (a) used to blur the image in Fig.~\ref{fig:qual}a.
        Deep-URL reconstructs the kernel with minimum shifts as compared to other iterative and deep learning methods (b-d).
        Classical RL algorithm completely fails and generates a noisy kernel (d).
        Note that \cite{nah2017deep} does not predict the motion blur kernel and hence is not included in the figure above.
        }
     \label{fig:kernel}
     \vspace{-0.4cm}
\end{figure}
\noindent \underline{\textbf{MNIST dataset results.}}
\label{exp:MNIST}
For this experiment, we consider the well-known MNIST dataset. We randomly draw 1000 sample images from the MNIST training dataset and use the same motion kernels provided by \cite{levin2009understanding}. 
Particularly, for each image, we convolve the original image with a randomly chosen aforementioned blur kernel to generate the degraded image.
Table~\ref{tab:MNIST} demonstrates the performances of the proposed Deep-URL framework with $L\in\{2,5\}$ layers and the original RL algorithm with the same number of iterations.
It is evident from Table~\ref{tab:MNIST} that the proposed method significantly outperforms the baseline RL algorithm across all evaluation metrics. 
Interestingly, Deep-URL achieves better performance in terms of both recovering the original image and the PSF even with only $L=2$ layers---this is presumably due to the hybrid model-based and data-driven nature of the proposed method.
Moreover, Deep-URL with $L=5$ layers attains a very high average ISNR value for the recovered image, which is 121$\times$ higher than that of the original RL algorithm. 
Note that, the RMSE between the original and the reconstructed PSF using the proposed method assumes a 8.55$\times$ smaller value than that of the RL algorithm. 
By comparing the evaluation performance of Deep-URL for $L\in\{2,5\}$, it is evident that increasing the number of layers result in a much higher increase of scores across all evaluation metrics as compared to the baseline RL algorithm.
Finally, from Fig.~\ref{fig:layer}, we found that the classical RL algorithm is sensitive to the number of iteration and the performance fluctuates on a random set of 100 MNIST images.
However, the performance of Deep-URL always increases as we increase the number of iterations \ie, the number of layers.\looseness=1

\noindent \underline{\textbf{Levin dataset results.}}
\label{exp:levin}
For this experiment, we use the dataset provided by \cite{levin2009understanding} -- a widely used benchmark dataset in several deblurring works
\cite{li2019algorithm, krishnan2011blind, xu2010two}.
It comprises of 4 grayscale images and 8 motion blur kernels: a total of 32 motion blurred images.
Table~\ref{tab:levin} summarizes the performance of Deep-URL in comparison with the baseline RL as well as the methodologies proposed in \cite{chakrabarti2016neural}, \cite{nah2017deep} and \cite{li2019algorithm} on the same dataset.
It can be observed from Table~\ref{tab:levin} that Deep-URL significantly outperforms the baseline RL algorithm across all image and kernel evaluation metrics.
In contrast to other methods that include \textit{a priori} learning using training images, Deep-URL is a self-deblurring framework and performs at par (PSNR) or better (ISNR and SSIM) on the image quality evaluation metrics.
Interestingly, an 1.8$\times$ increase can be observed in ISNR using Deep-URL with just $L=5$ when compared to \cite{li2019algorithm}.
In regards to the reconstructed blur kernel, it was found that most pixels did not converge to absolute zero and hence a higher RMSE score was obtained in reconstructing the motion kernel blindly.
From Fig.~\ref{fig:qual}, we observe Deep-URL reconstructs smoother images with lesser artifacts as compared to other state-of-the-art methods.
\setlength\tabcolsep{2pt}
\begin{table}[H]
\small
\centering
\vspace{-0.2cm}
\caption{
    \small
    Deep-URL (D-URL) outperforms RL algorithm (ran till 5 iterations) across all image quality and RMSE metrics.
    In contrast to existing deblurring methods which learn from training images, Deep-URL performs on par (PSNR) and better (ISNR and SSIM) in reconstructing the clean image.
}
\label{tab:levin}
\vspace{-0.2cm}
\begin{tabular}{@{}lrrrrrr@{}}
\toprule
Metrics & \cite{li2019algorithm} & \cite{nah2017deep} & \cite{chakrabarti2016neural}& RL & \makecell{D-URL \\($L=2$)} & \makecell{D-URL \\ ($L=5$)} \\ \midrule
PSNR(dB) & 27.15 & 24.51 & 23.18 & 19.42 & 24.85 & 27.12 \\
ISNR (dB) & 3.79 & 1.35 & 0.02& -2.98 & 5.36 & \textbf{6.95} \\
SSIM & 0.88 & 0.81 & 0.81 & 0.53 & 0.89 & \textbf{0.91} \\
RMSE($\times$1e-3) &
\textbf{3.87} & - & - & 10.10 & 8.08 &  7.10\\ \bottomrule
\end{tabular}
\vspace{-0.3cm}
\end{table}
\section{Conclusion}
\label{sec:conclusion}
In this work, we considered the problem of blind deconvolution and proposed the Deep-URL framework---a model-aware deep blind deconvolution architecture---by unfolding the Richardson-Lucy algorithm (Sec.~\ref{sec:method}).
Quantitative and qualitative evaluations (Sec.~\ref{sec:exp}) show Deep-URL achieves superior performance than both its baseline RL algorithm and several existing blind deconvolution techniques.
In contrast to other MAP-based frameworks, Deep-URL does not show convergence to the trivial solution of an impulse like kernel.
\bibliographystyle{IEEEbib}
\bibliography{refs}
\end{document}